\documentclass[11pt,a4paper]{llncs}
\usepackage{amsmath}
\usepackage{amssymb}
\usepackage{graphicx}
\usepackage{marvosym}
\usepackage{url}
\usepackage{fancyhdr}
\usepackage{multirow}

\usepackage{geometry}
\newcommand{\keywords}[1]{\par\addvspace\baselineskip
\noindent\keywordname\enspace\ignorespaces#1}

\fancypagestyle{firstpage}{\fancyhf{}
}

\begin{document}


\title{\LARGE{Dealing with negative samples with multi-task learning on span-based joint entity-relation extraction}}


%
%
\author{\large{Jiamin Lu \and Chenguang Xue}}
\institute{\large{College of Computer and Information, Hohai University, Nanjing, China, 
    \\ 211307050013@hhu.edu.cn}}

%


%
%


\maketitle

\thispagestyle{firstpage}

\begin{abstract}
Recent span-based joint extraction models have demonstrated significant advantages in both entity recognition and relation extraction. These models treat text spans as candidate entities, and span pairs as candidate relationship tuples, achieving state-of-the-art results on datasets like ADE. However, these models encounter a significant number of non-entity spans or irrelevant span pairs during the tasks, impairing model performance significantly. To address this issue, this paper introduces a span-based multitask entity-relation joint extraction model. This approach employs the multitask learning to alleviate the impact of negative samples on entity and relation classifiers. Additionally, we leverage the Intersection over Union(IoU) concept to introduce the positional information into the entity classifier, achieving a span boundary detection. Furthermore, by incorporating the entity Logits predicted by the entity classifier into the embedded representation of entity pairs, the semantic input for the relation classifier is enriched. Experimental results demonstrate that our proposed SpERT.MT model can effectively mitigate the adverse effects of excessive negative samples on the model performance. Furthermore, the model demonstrated commendable F1 scores of 73.61\%, 53.72\%, and 83.72\% on three widely employed public datasets, namely CoNLL04, SciERC, and ADE, respectively.
\end{abstract}


\section{Introduction}

The task of joint entity and relation extraction involves extracting the entities and their semantic relationships from unstructured textual data. This task forms the foundational basis for numerous downstream NLP tasks such as sentiment analysis and knowledge graph construction. Due to the limitations of traditional pipeline methods that overlook interactions between Named Entity Recognition (NER) and Relation Extraction (RE) tasks and suffer from error propagation, recent research has explored approaches for joint extraction\cite{eberts2019span}\cite{lai2021joint}\cite{ye2021packed}. Based on the classification strategy employed for extraction, existing models can be categorized into two types: sequence labeling-based models and span-based models. The former rely on sequence labeling mechanisms \cite{li2021joint}\cite{ma2022joint} to perform label-level classification. The latter, on the other hand, adhere to the span-based paradigm \cite{eberts2019span}\cite{tang2022boundary}\cite{ye2021packed} and engage in span-level classification.

Since the named entities contain nested entities, the traditional sequence labeling models face limitations on assigning multiple labels to a single token simultaneously. On the other hand, span-based models treat consecutive tokens as spans. For instance, in Figure 1, "Hypersensitivity," "Hypersensitivity to," and "Hypersensitivity to aspirin" represent spans of different lengths. Span-based models enumerate all possible spans and classify them into the correct categories, thereby eliminating the need for extensive feature engineering and effectively addressing the nested entity issue. However, these models encounter numerous challenges.

As the span strategy involves an enumeration process where each span (pair) needs to be classified, these models face an extremely imbalanced dataset. The dataset consists of significantly more non-entity spans (pairs) compared to entity spans (pairs). In the following text, we will refer to entity spans (pairs) as positive samples and non-entity spans (pairs) as negative samples. For example, in Figure 1, there are two entity spans, E1 and E2 (positive entity samples), and one relation R (positive relation sample). Instead , the span strategy continuously expands a sliding window to select all possible consecutive tokens, producing 64 negative entity samples. The pairwise combination of spans generates 4355 negative relation samples. Hence, it is evident that there exists a substantial disparity in the ratio between positive and negative samples. Previous span-based models did not take into account the adverse impact of excessive negative samples on model performance.

In the context of numerous negative samples, prior research in joint entity-relation extraction has employed the inclusion of a special entity class labeled as "NA" to classify negative samples. However, this approach disregards the recognition steps of Named Entity Recognition (NER) and Relation Extraction (RE) tasks, thus affecting the overall performance of the model. Nguyen et al. \cite{nguyen2015relation} proposed a recognition framework that breaks down RE into identification and classification stages, validating the efficacy of this approach for handling irrelevant entity pairs in RE. Our model draws inspiration from Nguyen et al.'s ideas, applying a multi-task framework to joint extraction and conducting a two-stage joint training for NER and RE.

Another challenge posed by the span-based method is the issue of hard negative samples. For instance, as shown in Figure 1, adopting the span strategy would lead to generating negative samples like "be manifested" and "a systemic anaphylactoid reaction". The former does not overlap with any entity, while the latter overlaps with the entity "systemic anaphylactoid reaction". The second type of span is termed hard negative samples. While the model can easily distinguish the first kind of negative sample, distinguishing the second kind is difficult. This challenge has been acknowledged by \cite{zheng2019boundary}\cite{cao2020incorporating}\cite{wei2019novel}, who introduced token-level boundary detection and boundary regression to address this problem. In our approach, we imply the Intersection over Union (IoU) within the multi-task framework to analyze hard negative samples quantitatively. In Nguyen et al.'s work \cite{nguyen2015relation}, they embedded label information from the NER task into the representation of entity pairs to enrich the information. Similarly, we found that the Logits from NER, based on the span-based strategy, also contain substantial semantic information. Consequently, in the RE stage, we embed the NER Logits information into the representation of span pairs to enhance joint extraction. In conclusion, our contributions are as follows:

\begin{itemize}
    \item We apply a multi-task learning approach to entity-relation joint extraction. For both entity recognition and relation extraction tasks, we employ a multi-task framework to separately identify and classify candidate entities/entity pairs, aiming to mitigate the impact of negative samples on model performance.
    \item We introduce the Intersection over Union (IoU) computation method into the entity recognition subtask. By leveraging IoU, we incorporate positional information to address the challenge of hard negative samples.
    \item We merge the entity Logits generated from the entity recognition task with the embeddings of span pairs, thereby enriching the input information for relation extraction.
\end{itemize}

The rest of this paper is organized as follows. Section 2 presents a review of relevant literature on joint extraction. Section 3 provides a comprehensive description of the proposed SpERT.MT model. Subsequently, Section 4 outlines the experiments, encompassing experimental setups, results, and a detailed analysis. Finally, in Section 5, we summarize our conclusions and identify potential directions for future research.

   \begin{figure} [ht]
   \begin{center}
   \begin{tabular}{c} 
   \includegraphics[width=0.8\linewidth]{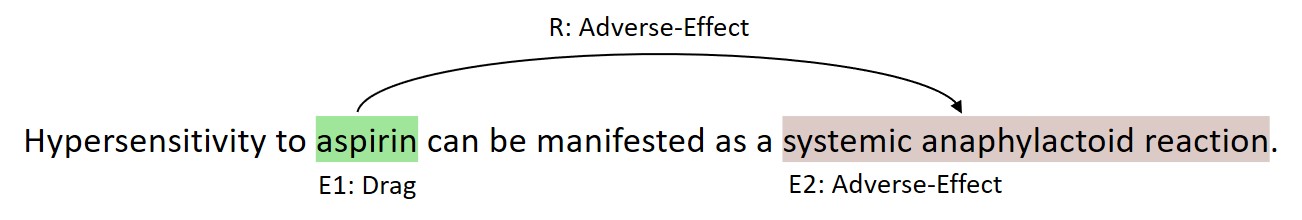}
   \end{tabular}
   \end{center}
   \caption[example] 
   { \label{fig:example} 
An example from the ADE dataset, wherein the entities E1 and E2 are delineated by shaded annotations. Employing the span-based strategy, these entities correspond to spans of length 1 and 3, respectively. The arrows symbolize the relationship R between E1 and E2."}
   \end{figure}

\section{Related work}

Joint entity and relation extraction encompass two subtasks: Named Entity Recognition (NER) and Relation Extraction (RE). Due to the error propagation and neglect of interactions between these subtasks, traditional pipeline methods, in recent years, have led to a growing interest in joint extraction \cite{lai2021joint}\cite{li2021joint}\cite{ma2022joint}\cite{tang2022boundary}\cite{ye2021packed}. Previous research can be classified into two strategies: token tagging and span classification.

The token tagging strategy treats extraction as a sequence labeling task \cite{li2021joint}\cite{ma2022joint}, where each token is labeled according to the BIO (or its variant BILOU) scheme. Miwa and Sasaki \cite{miwa2014modeling} approached joint entity and relation extraction as a table filling problem, employing a history-based beam search to find optimal table filling solutions. Ma et al. \cite{ma2022joint} also treated joint extraction as a table filling problem. In contrast to Miwa and Sasaki \cite{miwa2014modeling}, they incorporated the stacked model concept proposed by Miwa and Bansal \cite{miwa2016end}, stacking CNN on top of pre-trained language models to capture local dependencies and predict cell labels. Li et al. \cite{li2021joint} introduced a two-stage tagging scheme, labeling head entities and multiple tail entities separately for specific relationships. However, due to the limitation that a token cannot have multiple labels, these models cannot handle overlapping entities and relations. To address this, Wang et al. \cite{wang2020advanced} and Wei et al. \cite{wei2019novel} adopted multi-turn tagging, significantly increasing the time complexity for both training and decoding processes.

The span classification strategy involves naming entity classification for spans (subsequences of text) and relation extraction for all possible span pairs. Models adopting the span classification strategy usually perform exhaustive search over all spans, allowing them to detect nested entities. Luan et al. \cite{luan2018multi} introduced the first span-based model and conducted beam search over the hypothesis space. Their subsequent model, DyGIE \cite{luan2019general}, employed dynamically constructed span graphs to share span representations. Wadden et al. \cite{wadden2019entity} replaced the BiLSTM encoder with BERT, showcasing the advantages of using pre-trained models as encoders. Eberts and Ulges \cite{eberts2019span} encoded span representations using pre-trained models and utilized a strong negative sampling strategy to enhance the training process. Ye et al. \cite{ye2021packed} introduced the concept of pack-and-float tags, combining token tagging and span classification strategies to focus on relationships between spans/pairs. Tang et al. \cite{tang2022boundary} introduced boundary regression models to learn the offset between spans and the gold entities.

However, the aforementioned studies did not focus on the data imbalance issue caused by the span classification strategy. In contrast to their research, we address the data imbalance issue through a multi-task framework. Recently, numerous experiments have employed multi-task learning methods to train NER and RE models. Zheng et al. \cite{zheng2019boundary}, Cao et al. \cite{cao2020incorporating}, and Tan et al. \cite{tan2020boundary} used multi-task learning to capture the dependencies between entity boundaries and their classification labels. Fei et al. \cite{fei2019recognizing} proposed treating each unique entity type as a separate task to facilitate information exchange between biomedical NER tasks using multi-task learning. Similarly, Nguyen et al. \cite{nguyen2015relation} decomposed the RE task into two subtasks and employed multi-task learning to address the relation extraction issue in imbalanced corpora. The mentioned works collectively illustrate the favorable performance of multi-task learning methods in both NER and RE tasks.

The SpERT model, proposed by Eberts et al. \cite{eberts2019span}, is a span-based joint entity-relation extraction model. This model employs shallow classifiers for both NER and RE tasks and does not explicitly address the impact of data imbalance on the model. Our model is built upon the foundation of the SpERT model, and it utilizes a multi-task learning approach to individually optimize the NER and RE tasks. This is done to tackle the issue of training sample imbalance caused by the span classification strategy.

\section{Proposed Method}

The model proposed in this study, as illustrated in Figure 2, comprises four distinct modules: Vector Representation, Span Representation, Entity Recognition, and Relation Extraction. Given an input sentence, fine-tuned BERT is employed to generate token and contextual vector representations. Subsequently, the Span Representation layer enumerates all feasible spans, combining the outputs from the vector representation layer and applying width constraints to characterize these spans. Ultimately, the joint entity-relation extraction task is conducted based on the span representations. As this paper integrates a multi-task framework for both entity recognition and relation extraction, subsequent sections, apart from describing the four-layer structure of the SpERT model, will encompass the elucidation of the multi-task framework in Section 3.3, depiction of joint training loss in Section 3.6, and explication of the prediction methodology in Section 3.7.

   \begin{figure} [ht]
   \begin{center}
   \begin{tabular}{c} 
   \includegraphics[width=14.65cm, height=8.29cm]{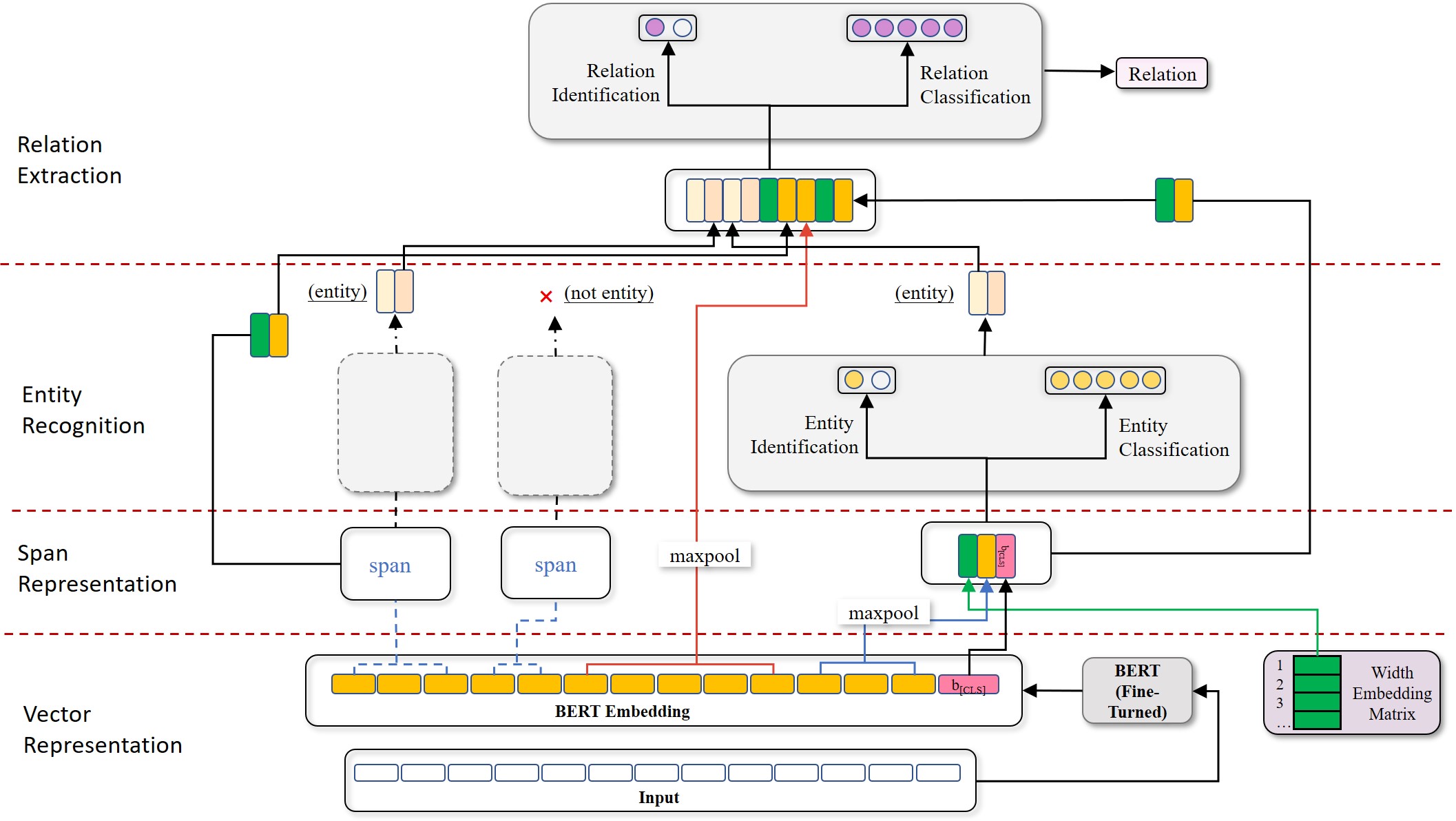}
	\end{tabular}
	\end{center}
   \caption[example] 
   { \label{fig:sample} 
Architecture of our proposed model, SpERT.MT.}
   \end{figure} 

\subsection{Vector Representation}
The Vector Representation layer employs fine-tuned pre-trained BERT to extract contextual information. As illustrated in the figure, the input sentence undergoes Byte Pair Encoding (BPE) and is split into a sequence of n tokens. BPE encoding represents uncommon words (e.g., "pretrain") with common subwords (e.g., "pre" and "train"), thus constraining the vocabulary and effectively handling out-of-vocabulary (OOV) and rare words. The BPE tokens are propagated through the internals of BERT, resulting in an embedded sequence E of length n+1: E = ($e_{[CLS]}$, $e_1$, $e_2$, …, $e_n$), where $e_{[CLS]}$ represents the embedded representation of the entire sentence context, and each embedding vector $\mathrm{e}_{\mathrm{i}}\in\mathbb{R}^{d_{1}}$ (with d1 denoting the embedding dimension).

\subsection{Span Representation}
The Span Representation layer utilizes the embeddings e of the tokens produced by BERT to represent spans. As illustrated in Figure 2, in order to maintain a consistent representation dimension for different spans s := ($e_i$, $e_{i+1}$, …, $e_j$), we apply a fusion process to s. We have chosen to employ max pooling for the fusion process:
\begin{equation}
    \mathbf{v}(s)=\text{maxpool}(\mathbf{e}_i,\mathbf{e}_{i+1},...,\mathbf{e}_j)\in\mathbb{R}^{d_1}
\end{equation}

Based on prior knowledge, it is unlikely that longer spans effectively represent meaningful entities. Therefore, we set a maximum span length L for the model, primarily to reduce complexity. In addition, we incorporate width information w ($0\mathrm{\leqslant w}\mathrm{\leqslant L}$) into the fused span representation s using a width embedding matrix. This approach is guided by the understanding that spans of varying widths contribute distinctively to the representation:
\begin{equation}
    \mathbf{c}(s)=\mathbf{v}(s_{ij})\circ\mathbf{w}_k\in\mathbb{R}^{d_1+d_2}
\end{equation}
Here, $\circ$ denotes the concatenation of vectors, and $w_k$ is derived from the width embedding matrix obtained through training the model.

The categories of entities within a sentence are likely to be correlated with the overall contextual information of the sentence. For instance, contextual keywords like "passing through" or "flying to" can strongly indicate the entity class "location". Therefore, we augment the embedding information of the context [CLS] to the span representation as well:
\begin{equation}
    \mathbf{x}(s)=\mathbf{c}(s)\circ\mathbf{e}_{[\mathrm{CLS}]}\in\mathbb{R}^{2d_1+d_2}
\end{equation}

\subsection{Multi-task Framework}

\subsubsection{Identification Task}
For the identification task, it is employed to determine whether the current entity or entity pair is indeed an entity or whether a relationship exists. This is a standard binary classification task, thus we opt to use binary cross-entropy as the loss function, as defined below:
\begin{equation}
    \mathcal{H}(p,q)=-\operatorname{qlog}(p)-(1-q)\log(1-p)
\end{equation}
Here, q $\in$ $\{0, 1\}$ designates the true class of the span or span pair (0 represents negative sample, 1 represents positive sample), and p $\in$ [0, 1] is the estimated probability by the model for the label q = 1 class.

Therefore, the loss for the identification task is computed as follows:
\begin{equation}
    \mathrm{Loss}_1=\Sigma\mathcal{H}(p,q)
\end{equation}

\subsubsection{Classification Task}
To enhance diversity in the loss landscape of multi-task learning and prevent the two tasks from degenerating into a single task, we opt for the pairwise ranking loss function (as proposed by Santos et al.\cite{santos2015classifying}) as the loss function for classification tasks. This loss function shares the same task objective as cross-entropy loss but possesses a distinct optimization direction.

Given the span (pair) representation s and the set of entity or relation classes C, we compute the score for class label c $\in$ C using the dot product:
\begin{equation}
    \mathrm{score}_c=s^T[W^\mathrm{classes}]_c
\end{equation}
Where $W^{classes}$ is the matrix to be learned, and the number of columns hit is the number of classes. $W_c^{classes}$ is the column vector corresponding to class c, with dimensions equal to the dimensions of s.

For a span (pair), where $y^+$ is the correct label and $y^-$ is not, the $scores_{y^+}$ and $scores_{y^-}$ are defined as the scores for $y^+$ and $y^-$, respectively. Then, the rank loss can be calculated as follows:
\begin{equation}
    L^+=\log\left(1+\exp\left(\gamma(m^+-\mathrm{score}_{y^+})\right)\right)
\end{equation}
\begin{equation}
    L^-=\log\left(1+\exp\left(\gamma(m^-+\mathrm{score}_y\text{-})\right)\right)
\end{equation}
Where $\gamma$ is a scaling factor, $m^+$ and $m^-$ are margins. For non-entity classes or irrelevant entity pairs, only $L^-$ is computed to penalize incorrect predictions. In the training step, the ranking loss selects only the highest scoring incorrect class among all error classes. Then, we optimize the pairwise ranking loss:
\begin{equation}
    Loss_2=\sum(L^++L^-)
\end{equation}
Finally, the loss function of our multi-task framework is formulated as:
\begin{equation}
    Loss=\alpha\mathrm{Loss}_1+\beta\mathrm{Loss}_2
\end{equation}

\subsection{Entity Recognition}
As shown in Figure 2, the entity recognition layer directly utilizes the output from the span representation layer as input, and utilizes the multi-task framework to extract entity categories from the spans. In this section, adjustments are made based on the multi-task framework described in Section 3.3.

In the identification task of the entity recognition multi-task framework, we introduce the positional information between spans using the Intersection over Union (IoU) metric. Formally, the IoU calculation is as follows:
\begin{equation}
    \begin{aligned}
    \operatorname{IoU}(s(i_1,j_1),s(i_2,j_2))&=
        \begin{cases}
            \frac{\min(j_1,j_2)-\max(i_1,i_2)+1}{\max(j_1,j_2)-\min(i_1,i_2)+1},&\min(j_1,j_2)\geqslant\max(i_1,i_2)
            \\0,&\mathrm{otherwise}
        \end{cases}
    \end{aligned}
\end{equation}
Where s(i,j) represents the span with left and right indices i and j in the sentence.

We compute and utilize the maximum IoU (Intersection over Union) between the span s and all entity spans in the sentence, defined as ENIoU(s). Its calculation formula is as follows:
\begin{equation}
    \mathrm{ENIoU(s)}=\max(\mathrm{IoU(s,en),en\in E})
\end{equation}
Where E represents the set composed of all entity spans in the sentence.

Subsequently, ENIoU is employed as a scaling factor to replace the loss function of the Identification task in the multi-task framework with dynamically scaled cross-entropy loss. The loss function is as follows:
\begin{equation}
    \mathcal{H}(p,q)=-q(1-\delta)\log(p)-\delta(1-q)(1+\text{ENIoU})\log(1-p)
\end{equation}
Where $\delta$ and ENIoU determine the scaling factor. $\delta \in [0,1]$ is a balancing factor to address class imbalance, and we set $\delta<{0.5}$ to emphasize the fewer positive samples. For negative samples, a higher ENIoU results in a larger loss, thereby focusing the model more on hard negative samples during training.

For the other parts of the multi-task framework, we do not make any further modifications in the entity recognition stage, and we denote the loss in the entity recognition stage as L1.

\subsection{Relation Extraction}
Unlike the direct use of span representation layer outputs as input in the entity recognition stage, the relation extraction stage makes some adjustments to the input information. For those spans that are classified as "NA" during the entity recognition multi-task stage, they will be filtered out. We only consider candidate span pairs that have been preliminarily identified as entities. As shown in Figure 2, we choose the local information between candidate span pairs as their context and obtain context representations by utilizing max-pooling to fuse their BERT embeddings.
\begin{equation}
    \mathbf{v}(s_1,s_2)=\text{maxpool}(\mathbf{e}_i,\cdots,\mathbf{e}_j)\in\mathbb{R}^{d_1+d_2}
\end{equation}
For adjacent candidate entity pairs, we set v($s_1$, $s_2$) = 0. Additionally, we observe that the Logits from the entity recognition stage may contain useful semantic information. Therefore, we further embed the Logits from the entity recognition stage to enrich the input information. Ultimately, the multi-task input representation of the relation extraction layer is as follows:
\begin{equation}
    \mathbf{r}(s_1,s_2)=\mathbf{c}(s_1)\circ\mathbf{v}(s_1,s_2)\circ\mathbf{c}(s_2)\circ\mathbf{p}(s_1)\circ\mathbf{p}(s_2)\in\mathbb{R}^{3d_1+2d_2+2d_3}
\end{equation}
Where p(s) represents the Logits from the entity recognition stage. Considering that the relationship between $s_1$ and $s_2$ is typically not symmetric, therefore,
\begin{equation}
    \mathbf{r}(s_2,s_1)=\mathbf{c}(s_2)\circ\mathbf{v}(s_1,s_2)\circ\mathbf{c}(s_1)\circ\mathbf{p}(s_2)\circ\mathbf{p}(s_1)\in\mathbb{R}^{3d_1+2d_2+2d_3}
\end{equation}

For the multi-task framework used in the relation extraction stage, it remains consistent with the approach described in Section 3.3. We denote the loss for the relation extraction stage as L2.

\subsection{Joint Training Loss}
We learn the parameters of width embedding w for spans (as shown in Figure 2) and the parameters of the entity/relation multi-task classifiers while fine-tuning BERT in this process. Due to our joint training approach for entity recognition and relation extraction, the loss is defined as follows:
\begin{equation}
    \mathbf{L}=L_1+L_2
\end{equation}

\subsection{Prediction}
At the prediction stage, to avoid error propagation, only the class scores obtained from the multi-task framework, denoted as $score_c$, are utilized, while the binary identification tasks are solely employed for optimizing network parameters.

Given a span(pair), the computation of the predicted class probability P is as follows:
\begin{equation}
    P=\left\{\begin{array}{cl}\text{argmax}_c(\text{score}_c), & \text{max}\left(\text{score}_c\right)\geqslant\theta \\ \text{NA}, & \text{max}\left(\text{score}_c\right)<\theta\end{array}\right.
\end{equation}
Where $\theta$ is a hyperparameter threshold. If the score for each class is below $\theta$, the span(pair) is predicted as non-entity or irrelevant relationship. Otherwise, the span(pair) is assigned to the class with the highest score.

\section{Experiments}
\subsection{Datasets}
\begin{itemize}
    \item CoNLL04: The CoNLL04 dataset \cite{roth2004linear} consists of sentences with annotated named entities and relationships extracted from news articles. The dataset comprises four entity types (People, Location, Organization, Other) and five relationship types (Live-In, Work-For, Kill, Located-In, Organization-Based-In). We adopt the split strategy proposed by Gupta et al. \cite{gupta2016table} for training (1153 sentences) and testing (288 sentences) the model. Additionally, we allocate 20\% of the training set for hyperparameter tuning.
    \item SciERC: The SciERC dataset \cite{luan2018multi} is derived from abstracts of 500 artificial intelligence research papers. It comprises six scientific entity types (Task, Method, Metric, Material, Other-Scientific-Term, Generic) and seven relationship types (Compare, Conjunction, Evaluate-For, Used-For, Feature Of, Part-Of, Hyponym-Of), totaling 2687 sentences. We adopt the same split strategy as described in \cite{luan2018multi} for training (1861 sentences), development (275 sentences), and testing (551 sentences) on the SciERC dataset.
    \item ADE Dataset: The ADE dataset \cite{gurulingappa2012development} comprises 4272 sentences and 6821 relationships extracted from medical reports that describe adverse reactions caused by drug usage. The dataset encompasses two entity types (Adverse-Effect, Drug) and a single relationship type (Adverse-Effect). Due to the absence of official splits, we follow the approach of previous studies and conduct a 10-fold cross-validation for evaluation.
\end{itemize}

\subsection{Implementation}
We employed BERT as the encoder for the CoNLL04 dataset, while for ADE and SciERC, we utilized BioBERT and SciBERT with domain-specific features, respectively. As SpERT \cite{eberts2019span} served as our experimental baseline, we adopted some of its hyperparameter configurations. Specifically, we initialized the classifier weights with normally distributed random numbers ($\mu$=0, $\sigma$=0.02). The Adam optimizer was employed for model training, featuring linear warm-up and linear decay, with a peak learning rate set at 5$e^{-5}$. The batch size was set to 2, and the width embedding dimension of matrix W was 25.

For the hyperparameters in the multi-task framework, we utilized the same settings for entity recognition and relation extraction in the pairwise ranking loss: $m^+$ was set to 2.5, $m^-$ to 0.5, and $\gamma$ to 2. For the dynamic scaling of the cross-entropy loss in the identification phase of entity recognition, $\sigma$ was set to 0.25 ($\sigma$<0.5). In the prediction phase, the threshold $\theta$ was set to 0.85.

Across both entity recognition and relation extraction, we set the number of negative samples to 120 and performed training for each dataset over 20 epochs. Importantly, we employed the same set of hyperparameters for all three datasets, refraining from dataset-specific adjustments.

\subsection{Evaluation Metrics}
In the entity recognition phase, an entity is considered correct if the predicted entity type and span match the given textual span. For relation extraction, we adopted an evaluation method consistent with that proposed in \cite{eberts2019span}. Specifically, a relation is deemed correct if the relationship type between the provided span pairs and the entities represented by the spans are accurate. This evaluation criterion, in contrast to strict relation extraction, does not require verifying the correctness of the entity types associated with the spans, thus alleviating the need to focus on the results of Named Entity Recognition NER.

We evaluate the model performance on different datasets using various metrics including precision (both micro and macro), recall, and F1 score.

\subsection{Overall Results}
The table 1 presents a performance comparison between our model and several of the previous state-of-the-art entity-relation joint extraction models on the test datasets. Due to substantial variability in our model's measurements, as also indicated in the study by Taillé et al. \cite{taille2020let}, we report the average of 10 runs for each dataset except ADE. Regarding the choice of precision evaluation, different standards are applied to different datasets: for the CoNLL04 and SciERC datasets, we use micro-averaged F1 score; for the ADE dataset, we employ macro-averaged F1 score. Additionally, for the ADE dataset, the reported results account for experiments allowing overlapping entities.

From the table 1, it is evident that our model demonstrates significant performance improvements compared to the baseline model SpERT. Moreover, our model achieves competitive results when compared to the state-of-the-art models. Particularly noteworthy is the remarkable performance achieved on the SciERC dataset, where our model attains the state-of-the-art experimental results. Specifically:
\begin{itemize}
    \item CoNLL04: In the NER task, our model outperforms the SpERT baseline by an increase of 1.76\% in F1 score. In the RE task, our model achieves a 2.14\% higher F1 score compared to SpERT. When compared to state-of-the-art models, our model exhibits a notable advantage in NER due to the incorporation of IoU-enhanced span boundary detection. However, in the RE task, while we enhance input semantics, we did not specifically optimize for the RE task, leading to a noticeable gap from the state-of-the-art performance.
    \item SciERC: On the SciERC dataset, our model achieves state-of-the-art performance with F1 scores of 73.22\% in NER and 53.72\% in RE tasks. This success can be attributed to our choice of using the SciBERT model, which is specifically tailored for domain-specific encoding. Compared to several other state-of-the-art models, our experimental results surpass the state-of-the-art in all metrics except precision in the RE task, where we fall slightly behind.
    \item ADE: Our model also exhibits impressive performance on the ADE dataset. Compared to the SpERT baseline, we achieve improvements of 2.46\% and 4.88\% in NER and RE F1 scores, respectively. Under the condition of excluding the experimental results of Tang et al., our model achieves state-of-the-art performance. The data in the table illustrates that within this cohort of state-of-the-art models, our model excels in NER, surpassing RE performance, and particularly excels in NER recall, achieving state-of-the-art results.
\end{itemize}

\begin{table}[ht]
    \caption{Comparison between existing methods and our proposed SpERT.MT model on the CoNLL04, SciERC, and ADE datasets. The symbols $\vartriangle$ and $\blacktriangle$ represent evaluation using micro and macro average F1 values, respectively.} 
    \label{tab:Comparison}
    \begin{center}       
        \begin{tabular}{|l|l|l|l|l|l|l|l|}
            \hline
            \rule[-1ex]{0pt}{3.5ex}  \multirow{2}*{Dataset} & \multirow{2}*{\centering Model} & \multicolumn{3}{|c|}{NER} & \multicolumn{3}{|c|}{RE}  \\
            \cline{3-8}
            \rule[-1ex]{0pt}{3.5ex}  ~ & ~ & Precision & Recall & F1 & Precision & Recall & F1    \\
            \hline
            \rule[-1ex]{0pt}{3.5ex}  \multirow{5}*[-2ex]{CoNLL04$\vartriangle$} & SpERT(2020) & 88.25 & 89.64  & 88.94 & 73.04 & 70.00  & 71.47 \\
            \cline{2-8}
            \rule[-1ex]{0pt}{3.5ex}    ~ & ECA Net(2021)      & \textbf{91.14}     & 80.80  & 86.79 & \textbf{82.42}     & 73.34  & \textbf{77.55} \\
            \cline{2-8}
            \rule[-1ex]{0pt}{3.5ex}    ~ & TablERT-CNN(2022)  & -         & -      & 90.50  & -         & -      & 73.20  \\
            \cline{2-8}
            \rule[-1ex]{0pt}{3.5ex}    ~ & Tang et al.(2022)  & 90.80     & 88.80  & 89.80 & 76.60     & \textbf{74.40}  & 75.50 \\
            \cline{2-8}
            \rule[-1ex]{0pt}{3.5ex}    ~ & SpERT.MT(Ours)                & 90.13     & \textbf{91.27}  & \textbf{90.70} & 75.36     & 71.94  & 73.61 \\
            \hline
           \rule[-1ex]{0pt}{3.5ex} \multirow{7}*[3ex]{SciERC$\vartriangle$}  & SpERT(2020)        & 70.87     & 69.79  & 70.33 & 53.40     & 48.54  & 50.84 \\
           \cline{2-8}
           \rule[-1ex]{0pt}{3.5ex} ~ & Tang et al.(2022)  & 62.40     & 67.10  & 64.70 & \textbf{56.60}     & 48.20  & 52.10 \\
           \cline{2-8}
           \rule[-1ex]{0pt}{3.5ex} ~ & PL-Marker(2022)    & -         & -      & 69.90  & -         & -      & 53.20  \\
           \cline{2-8}
           \rule[-1ex]{0pt}{3.5ex} ~ & SpERT.MT(Ours)     & \textbf{71.21}     & \textbf{75.35}  & \textbf{73.22} & 53.63     & \textbf{53.81}  & \textbf{53.72} \\
           \hline
           \rule[-1ex]{0pt}{3.5ex} \multirow{5}*[-2ex]{ADE$\blacktriangle$}     & SpERT(2020)        & 88.99     & 89.59  & 89.28 & 77.77     & 79.96  & 78.84 \\
           \cline{2-8}
           \rule[-1ex]{0pt}{3.5ex} ~ & KECI(2021)         & -         & -      & 90.67 & -         & -      & 81.74 \\
           \cline{2-8}
           \rule[-1ex]{0pt}{3.5ex} ~ & ECA Net(2021)      & 88.52     & 81.11  & 84.63 & 82.28     & 77.17  & 79.62 \\
           \cline{2-8}
           \rule[-1ex]{0pt}{3.5ex} ~ & TableERT-CNN(2022) & -         & -      & 89.70  & -         & -      & 80.50  \\
           \cline{2-8}
           \rule[-1ex]{0pt}{3.5ex} ~ & Tang et al.(2022)  & \textbf{93.60}      & 91.50   & \textbf{92.50}  & \textbf{85.40}      & 85.00   & \textbf{85.20}  \\
           \cline{2-8}
           \rule[-1ex]{0pt}{3.5ex} ~ & SpERT.MT(Ours)     & 91.35     & \textbf{92.77}  & 92.05 & 81.06     & \textbf{86.55}  & 83.72 \\
           \hline
        \end{tabular}
    \end{center}
\end{table} 

\subsection{Ablation Study}
We conducted ablation experiments on the SciERC dataset, and the data in the table validates the effectiveness of our Multi-Task Learning (MTL) joint learning approach, the incorporation of IoU-enhanced dynamic scaling Cross-Entropy Loss (EIL), and the utilization of entity Logits. Each category of data represents the average of three experimental results. Observing the results, we note that our Multi-Task Entity-Relation Joint Extraction approach improves the F1 scores for entities and relations by 1.07\% and 0.83\%, respectively. This suggests that transitioning from the original single-step multi-class approach to the binary and multi-class multi-task approach contributes to enhanced model performance. Removing the EIL from our model results in a decrease of 1.01\% and 0.61\% in F1 scores for entities and relations, respectively. This indicates that introducing the IoU-enhanced dynamic scaling Cross-Entropy Loss during the entity recognition phase significantly benefits entity identification compared to relation extraction. In the relation extraction phase, enriching the embedding representation with entity Logits leads to F1 score improvements of 0.94\% and 1.23\% for entities and relations, respectively. This highlights that enhancements made during the relation extraction phase in the context of joint training for entity and relation extraction also contribute to improved entity recognition performance.

\begin{table}[ht]
    \caption{Ablation study of SpERT.MT on SciERC.} 
    \label{tab:Paper Margins}
    \begin{center}       
        \begin{tabular}{|l|l|l|l|l|l|l|} 
            \hline
            \rule[-1ex]{0pt}{3.5ex}   \multirow{2}*{Model} & \multicolumn{3}{|c|}{NER} & \multicolumn{3}{|c|}{RE}  \\
            \cline{2-7}
            \rule[-1ex]{0pt}{3.5ex}  ~ & Precision & Recall & F1 & Precision & Recall & F1   \\
            \hline
            \rule[-1ex]{0pt}{3.5ex}  \textbf{SpERT.MT(SciBERT)} & \textbf{71.21} & \textbf{75.35}  & \textbf{73.22} & \textbf{53.63} & \textbf{53.81}  & \textbf{83.72}  \\
            \hline
            \rule[-1ex]{0pt}{3.5ex}  \textbf{-MTL} & 70.51 & 73.87  & 72.15 & 52.99 & 52.79  & 52.89  \\
            \hline
            \rule[-1ex]{0pt}{3.5ex}  \textbf{-Entity Logits} & 70.59 & 74.06  & 72.28 & 52.86 & 52.12  & 52.49  \\
            \hline 
            \rule[-1ex]{0pt}{3.5ex} \textbf{-EIL} & 70.86 & 73.62  & 72.21 & 53.13 & 53.09  & 53.11 \\
            \hline 
        \end{tabular}
    \end{center}
\end{table}

\subsection{Negative Sampling}
In our study, the hyperparameters of the number of negative samples and the maximum span length (W) can influence the selection of negative samples during training. For the maximum span length (W), we considered dataset characteristics and the hyperparameters of the SpERT model. Accordingly, we set W to 10. The focus of this experiment lies in investigating model performance (expressed through F1 scores) under scenarios with excessive negative samples in the context of a multi-task framework. We continually adjusted the hyperparameters for the number of negative samples for entities and relations to observe model performance. Taking cues from the SpERT model, we also considered the average sentence length in the dataset, setting a limit of 200 for the experiment's maximum value. For sentences with negative sample counts exceeding the maximum limit, we randomly sampled from all samples.

\begin{figure} [ht]
    \begin{center}
        \begin{tabular}{c} 
            \includegraphics[width=0.6\linewidth]{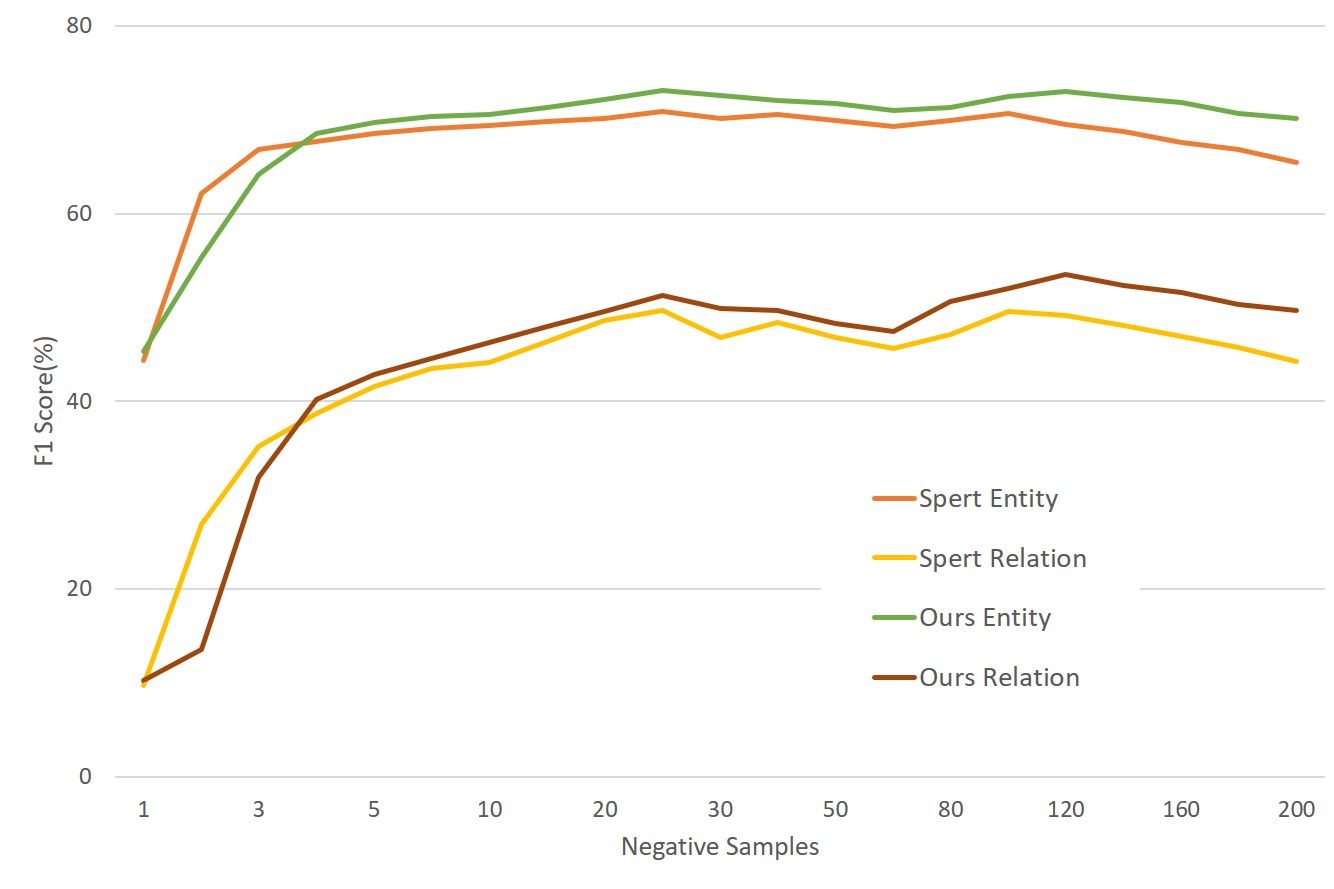}
        \end{tabular}
    \end{center}
    \caption[example] 
    { \label{fig:negative} 
    The Variation of F1 Scores for Entity and Relation Classification between Our Model and SpERT Model under Different Negative Sample Quantities.}
\end{figure} 

Figure 3 illustrates the F1 scores for entity recognition and relation extraction of our model and the SpERT model on the SciERC dataset with varying numbers of negative samples. As observed in the graph, when negative samples are scarce, model performance is poor. With increasing negative sample counts, performance improves significantly, in line with the conclusion from \cite{eberts2019span} that negative samples aid model training. Performance stabilizes as negative sample counts rise, yet when the count reaches 100-120, the model's performance generally peaks. Beyond this range, increasing negative samples can cause performance to decline, suggesting the need for a balance between an adequate number of negative samples and a limit on their maximum value. Our model outperforms the SpERT model when a certain number of negative samples are present. The peak performance of our model occurs later than that of SpERT when a sufficient number of negative samples are available, implying better performance for our model in multi-negative sample scenarios. Additionally, our model exhibits a gentler decline in performance gradient with further increases in negative sample counts compared to the SpERT model. These results indicate that the multi-task training framework is advantageous in scenarios with excessive negative samples leading to data imbalance. Based on the experimental data, we ultimately chose a maximum negative sample limit of 120 for both entities and relations.

\section{CONCLUSION}
This paper introduces a model, SpERT.MT, based on fine-tuning pretrained BERT and utilizing a span-based strategy for joint entity-relation extraction. To address the issue of excessive negative samples resulting from span-based strategies, we employ a multi-task framework for both entity recognition and relation extraction tasks. Considering the specific characteristics of entity recognition and relation extraction, we adjusted the loss functions and input embeddings within the multi-task framework accordingly. Through experiments involving different numbers of negative samples, we demonstrate that the multi-task framework improves model performance in scenarios with an abundance of negative samples. We observed that the tasks of joint entity-relation extraction are both built upon a common span representation. In light of this, future work will involve enriching the representation information at the span level to validate our hypothesis.

\section*{ACKNOWLEDGMENTS}       
 
The work is supported in part by the National Key R\&D Program of China (Grant No. 2021YFB3900605 \& 2021YFB3900601).

\vspace{2cm}

\section*{Authors}
\noindent {\bf Jiamin Lu} Assistant Professor at Information Department in Hohai University, China. He received his Ph.D degree in Information Science from FernUniversität in Hagen, Germany, 2014. His research interests include parallel processing on MOD (Moving Object Database), data management in Knowledge Graph construction.  At present, he is mainly working on the conjunction of big data technologies and smart water applications.\\

\noindent {\bf Chenguang Xue} Currently, he is pursuing his Master in Computer Science from the University of Hohai. His research interests include Natural Language Processing and Knowledge Graph Construction.\\

\end{document}